%% file: main.tex
\documentclass[11pt,a4paper]{article}
\usepackage[hyperref]{emnlp-ijcnlp-2019}
\usepackage{times}
\usepackage{latexsym}
\usepackage{booktabs}
\usepackage{multirow}
\usepackage{url}
\usepackage{algorithm}
\usepackage[noend]{algpseudocode}
\usepackage{url}

\usepackage{graphicx}
\graphicspath{ {./graphs/} }
\usepackage{tikz}
\usetikzlibrary{shapes, shapes.geometric, shapes.multipart, arrows, backgrounds, positioning, decorations.pathreplacing}
\usepackage{subcaption}

\aclfinalcopy

\input{mymacros.tex}

\title{Cross-lingual Parsing with Polyglot Training and Multi-treebank Learning: A Faroese Case Study}

\author{James Barry \and Joachim Wagner \and Jennifer Foster \\
  ADAPT Centre \\ 
  School of Computing,
  Dublin City University, Ireland \\
  {\tt firstname.lastname@adaptcentre.ie}
}

\date{}

\begin{document}
\maketitle

\begin{abstract}
\input{abstract.tex}

\end{abstract}

\section{Introduction}
\label{sec:intro}
\input{intro.tex}

\section{Background}
\label{sec:lit}
\input{background.tex}

\section{Method}
\label{sec:method}
\input{method.tex}

\section{Experiments}
In this section, we describe our experiments,
which include a replication of the main findings of \newcite{tyers-etal-2018-multi}, using AllenNLP \cite{gardner-etal-2018-allennlp} for POS tagging and parsing
instead of UDPipe
\cite{straka-strakova-2017-tokenizing}.\footnote{All of the code and scripts to reproduce the experiments can be found at \url{https://github.com/Jbar-ry/multilingual-parsing}}

\subsection{Details}
\input{model-details.tex}

\input{experiments.tex}
\subsection{Results}
\input{results_dev.tex}
\input{sentence-stats.tex}
\input{results_faroese.tex}

\input{results-sampled.tex}
\input{results-previous.tex}
\input{results.tex}

\section{Conclusion}
\label{sec:conclusion}
\input{conclusion.tex}

\section*{Acknowledgments}
\input{ack.tex}

\bibliographystyle{acl_natbib}
\bibliography{main}


\end{document}

%% file: mymacros.tex
\definecolor{DarkRed}{RGB}{140,5,0}

\newcommand{\ie}{i.\,e.\ }
\newcommand{\eg}{e.\,g.\ }
\newcommand{\vs}{vs.\ }

\newcommand{\naive}{na\"{\i}ve}

\newcommand{\bokmaal}{Bokm\r{a}l}

%% file: abstract.tex
Cross-lingual dependency parsing involves transferring syntactic knowledge from one language to another.
It is a crucial component for inducing dependency parsers in low-resource scenarios where no training data for a language exists.
Using Faroese as the target language, we compare two approaches using annotation projection: first, projecting from multiple \emph{monolingual} source models; second, projecting from a single \emph{polyglot} model which is trained on the combination of all source languages.
Furthermore, we reproduce multi-source projection \cite{tyers-etal-2018-multi}, in which dependency trees of multiple sources are combined.
Finally, we apply multi-treebank modelling to the projected treebanks, in addition to or alternatively to polyglot modelling on the source side.
We find that polyglot training on the source languages produces an overall trend of better results on the target language but the single best result for the target language is obtained by projecting from monolingual source parsing models and then training multi-treebank POS tagging and parsing models on the target side.

%% file: intro.tex
Cross-lingual transfer methods, \ie methods that
transfer
knowledge from one or more source languages to a
target language, have led to substantial improvements for low-resource 
dependency parsing \cite{rosa-marecek-2018-cuni, agic-etal-2016-multilingual, guo-etal-2015-cross,lynn-etal-2014-cross, mcdonald-etal-2011-multi, hwa2005bootstrapping} and part-of-speech (POS) tagging \cite{plank-agic-2018-distant}.
In low-resource scenarios,
there may be not enough data for data-driven models 
to learn how to parse.
In cases where no annotated data is available, knowledge is often transferred from annotated data in other languages and when there is only a small amount of annotated data, additional knowledge can be induced from external corpora such as by learning distributed word representations \cite{mikolov2013distributed, al-rfou-etal-2013-polyglot} and more recent contextualized variants \cite{peters-EtAl:2018:N18-1, devlin-etal-2019-bert}.

This work focuses on dependency parsing for low-resource languages by means of annotation projection \cite{yarowsky2001inducing} and synthetic treebank creation \citep{TiedemannZeljko16}.
We build on recent work by \newcite{tyers-etal-2018-multi} who show that in the absence of annotated training data for the target language, a lexicalized treebank can be created by translating a target language corpus into a number of related source languages and parsing the translations using models trained on the source language treebanks.\footnote{In
    this paper, \emph{source language}
    and \emph{target language} always refer to the projection, not the direction of translation.
}
These annotations are then projected to the target language using separate word alignments for each source language, combined into a single graph for each sentence and decoded 
\cite{sagae-lavie-2006-parser}, resulting in a treebank for the target language,
Faroese in the case of \citeauthor{tyers-etal-2018-multi}'s and our experiments.

Inspired by recent literature involving multilingual learning \cite{mulcaire-etal-2019-polyglot, smith-etal-2018-82, vilares-etal-2016-one},  
we investigate whether additional improvements can be made
by:
\begin{enumerate}
    \item using a single polyglot\footnote{We
              adopt the same terminology used in \newcite{mulcaire-etal-2019-polyglot}, who use the term \emph{cross-lingual transfer} to describe methods involving the use of one or more source languages to process a target language.
              They reserve the term \emph{polyglot learning}
              for training a single model on
              multiple languages and where parameters are shared between languages.
              For the polyglot learning technique applied to multiple treebanks
              of a single language, we use the term
              \emph{multi-treebank learning}.
          }
          parsing model
          which is trained on the combination of
          all source languages to create synthetic source treebanks
          (which are subsequently projected to the target language)
    \item training a multi-treebank
          model on the
          individually projected treebanks and the
          treebank produced with multi-source projections.
\end{enumerate}

The former differs from the approach of \newcite{tyers-etal-2018-multi}, who use multiple discrete, monolingual models to parse the translated sentences, whereas in this work we use a single model trained on multiple source treebanks.
The latter differs from training on the target treebank produced by multi-source projection in that the information of the individual projections is still available and training data is not reduced to cases where all source languages provide a projection.

In other words, 
we aim to investigate whether the current state-of-the-art approach for Faroese, which relies on cross-lingual transfer, can be improved upon by adopting an approach based on
source-side
polyglot learning
and/or target-side multi-treebank learning.
We hypothesize that a polyglot model can exploit similarities in morphology and syntax across the included source languages, which will result in a better model to provide annotations for projection.
On the target side, we expect that combining different sources of information will result in a more robust target model.

We evaluated our various models on the Faroese test set and experienced considerable gains for three of the four source languages (Danish, Norwegian \bokmaal{} and Swedish) by adopting a polyglot model.
However, for Norwegian Nynorsk, a stronger monolingual model was able to outperform the polyglot approach.
When we extended multi-treebank learning to the target side, we experienced additional gains for all cases. Our best result of 71.5 -- an absolute improvement of 7.2 points over the result reported by \newcite{tyers-etal-2018-multi} -- was achieved with multi-treebank target learning over the monolingual projections.

%% file: background.tex
\newcite{tyers-etal-2018-multi} describe a method for creating synthetic treebanks for Faroese based on previous work which uses machine translation and word alignments to transfer trees from source language(s) to the target language.
Sentences from Faroese are translated into the four source languages Danish, Swedish, Norwegian Nynorsk and Norwegian \bokmaal{}.
The translated sentences are then tokenized, POS tagged and parsed using the relevant source language model trained on the source language treebank.
The resulting trees are projected back to the Faroese sentences using word alignments.
The four trees for each sentence are combined into a graph with edge scores one to four (the number of trees that support them), from which a single tree per sentence is produced using the Chu-Liu-Edmonds algorithm \cite{chu1965shortest, edmonds1967optimum}.
The resulting trees make up a synthetic treebank for Faroese which is then used to train a Faroese parsing model.
The parser output is evaluated using the gold-standard Faroese test treebank developed by \newcite{tyers-etal-2018-multi}.
The approach is compared to a delexicalized baseline, which it outperforms by a large margin.
It is also shown that, for Faroese, a combination of the four source languages (multi-source projection) is superior to individual language projection.

The idea of annotation projection using word-alignments originates from \cite{yarowsky2001inducing} who used word alignments to transfer information such as POS tags from source to target languages.
This method was later used in dependency parsing by \newcite{hwa2005bootstrapping}, who project dependencies to a target language and use a set of heuristics to form dependency trees in the target language.
A parser is then trained on the projected treebank and evaluated against gold-standard treebanks.
\newcite{zeman-resnik-2008-cross} 
introduced the idea of delexicalized dependency parsing whereby a parser is trained using only POS information and is then applied to a target language.

\newcite{mcdonald-etal-2011-multi} perform delexicalized dependency parsing using direct transfer and show that this approach outperforms unsupervised approaches for grammar induction. Importantly, this approach can be extended to the multi-source case by training on multiple source languages and predicting a target language.
In an additional experiment, they combine annotation projection and multi-source transfer.

\newcite{TiedemannZeljko16} present a thorough comparison of pre-neural cross-lingual parsing. Various forms of projected annotation methods are compared to delexicalized baselines, and the use of machine translation instead of parallel corpora to produce synthetic treebanks in the target language is explored. In contrast to \newcite{tyers-etal-2018-multi}, they translate a target sentence and project the source parse tree  back to the target during test time instead of using this approach to obtain training data for the target language.

\newcite{agic-etal-2016-multilingual} leverage massively multilingual parallel corpora such as translations of the Bible and web-scraped data from the Watchtower Online Library website\footnote{\url{https://wol.jw.org/}} for low-resource POS tagging and dependency parsing using annotation projection.
They project weight matrices (as opposed to decoded dependency arcs) %
from multiple source languages and average the matrices weighted by word alignment confidences.
They then decode the weight matrices into dependency trees on the target side, which are then used to train a parser.
This approach utilizes dense information from multiple source languages, which helps reduce noise from source side predictions but to the best of our knowledge, the source-side parsing models
learn information between source languages independently and the cross-lingual interaction only occurs when projecting the edge scores into multi-source weight matrices.

The idea of projecting dense information in the form of a weighted graph has been further extended by \newcite{schlichtkrull-sogaard-2017-cross} who bypass the need to train the target parser on decoded trees and develop a parser which can be trained directly on weighted graphs.

\newcite{plank-agic-2018-distant} use annotation projection for POS tagging.
They find that choosing high quality training instances results in superior accuracy  than randomly sampling a larger training set.
To this end, they rank the target sentences by the percentage of words covered by word alignments across all source languages and choose the top $k$ covered sentences for training.

\newcite{meechanmadden2019} carry out an evaluation on cross-lingual parsing for three low-resource languages which are supported by related languages. They include three experiments: first, training a \emph{monolingual} model on a small number of sentences in the target language; second, training a \emph{cross-lingual} model on related source languages which is then applied to the target data and lastly, training a \emph{multilingual} model which includes target data as well as data from the related support languages.
They found that training a monolingual model on the target data was always superior to training a cross-lingual model.
Interestingly, they found that the best results were achieved by training a model on the various support languages as well as the target data, \ie their multilingual model. While we do not combine the synthetic target treebanks with the source treebanks in our experiments, the results of \newcite{meechanmadden2019} motivate us to carry out this experiment in the future.

%% file: method.tex
We outline the process used for creating a synthetic treebank for cross-lingual dependency parsing.
We use the following resources:
raw Faroese sentences taken from Wikipedia,
a machine translation system to translate these sentences into all source languages (Danish, Swedish, Norwegian Nynorsk and Norwegian \bokmaal{}),
a word-aligner to provide word alignments between the words in the target and source sentences,
treebanks for the four source languages on which to train parsing models,
%
POS tagging and parsing tools, and,
lastly a target language test set. %
We use the same raw corpus, alignments and tokenized and segmented versions of the source translations\footnote{The original authors tokenize and segment the source translations with UDPipe.} as \newcite{tyers-etal-2018-multi} who release all of their data.\footnote{\url{https://github.com/ftyers/cross-lingual-parsing}}
In this way, the experimental pipeline is the same as theirs but we predict POS tags and dependency annotations using our own models.

\paragraph{Target Language Corpus}

We use the target corpus built by \newcite{tyers-etal-2018-multi} which comprises 28,862 sentences which were extracted from Faroese Wikipedia dumps\footnote{\url{https://dumps.wikimedia.org/}} using
 the WikiExtractor script\footnote{\url{https://github.com/attardi/wikiextractor}} and further pre-processed 
 to remove any non-Faroese texts and other forms of unsuitable sentences.

\paragraph{Machine Translation}
\label{sec:data:mt}

\begin{figure}[htb]
    \centering\includegraphics[width=7.7cm]{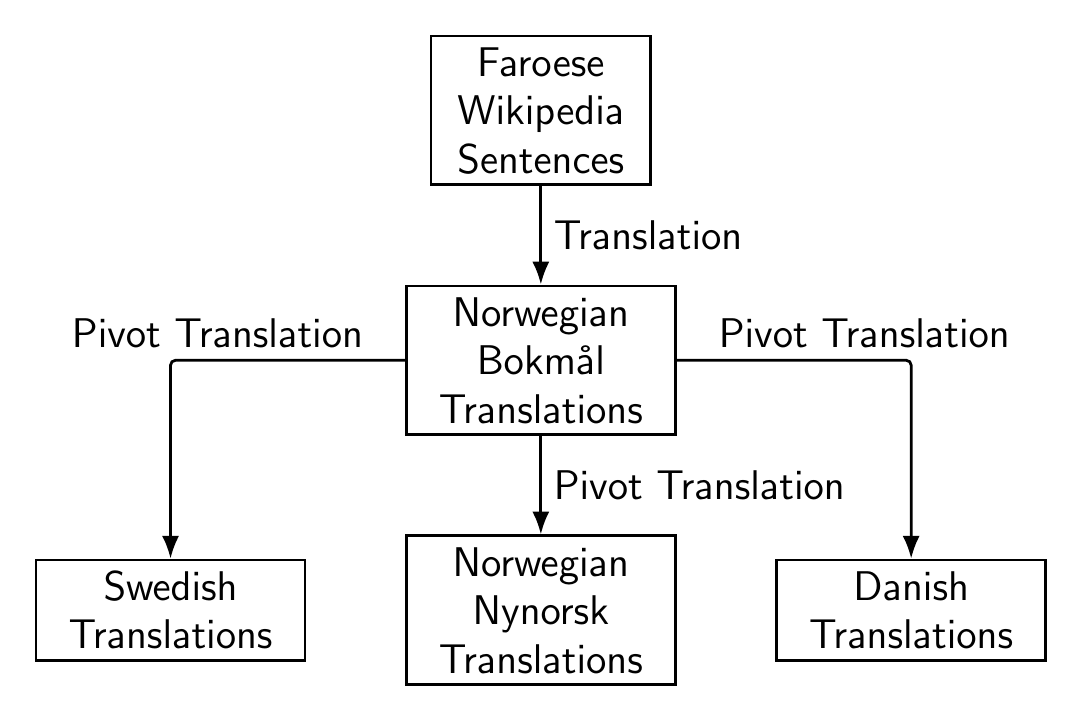}
    \caption{Overview of the machine translation process. The Faroese sentences are first translated into Norwegian \bokmaal{} and then from Norwegian \bokmaal{} into the other source languages (pivot translation).}
  \label{fig:translation}
\end{figure} 

As noted by \newcite{tyers-etal-2018-multi}, popular repositories for developing machine translation systems such as OPUS \cite{tiedemann2016opus} contain an inadequate amount of sentences to train a data-driven machine translation system for Faroese.
For instance, there are fewer than 7,000 sentence pairs between Faroese and Danish, Faroese and English, Faroese and Norwegian and Faroese and Swedish.
Consequently, to create parallel source sentences, \newcite{tyers-etal-2018-multi} use a rule-based machine translation system available in 
Apertium\footnote{\url{https://github.com/apertium}}
to translate from Faroese to Norwegian \bokmaal{}.
There also exists translation systems from Norwegian \bokmaal{} to Norwegian Nynorsk, Swedish and Danish in Apertium.
As a result, the authors use \emph{pivot translation} from Norwegian \bokmaal{} into the other source languages. The process is illustrated in Fig.~\ref{fig:translation}.
For a more thorough description of the machine translation process and for resource creation in general, see the
work of \newcite{tyers-etal-2018-multi}.

\paragraph{Word Alignments}
\label{sec:method:alignment}

We use word alignments between the Faroese text and the source translations generated  by \citet{tyers-etal-2018-multi} using fast\textunderscore{align} \cite{dyer-etal-2013-simple}, a word alignment tool based on IBM Model 2.\footnote{Note that previous related work \cite{agic-etal-2016-multilingual} report better results using IBM Model 1 with a more diverse language setup. They claim that IBM Model 2 introduces a bias towards more closely related languages.
As we are working with related languages and translations and alignments are largely word-for-word, we expect that this will have less of an impact on our experiments although IBM Model 1 should also be tried in future work.}


\paragraph{Source Treebanks}
\label{sec:method:treebanks:authentic}

We use the Universal Dependencies v2.2 treebanks \cite{11234/1-2837} to train our source parsing models.
This is the version used for the 2018 CoNLL shared task on Parsing Universal Dependencies~\cite{zeman:EtAl:2018:K18:2}.

\paragraph{Source Tagging and Parsing Models}
\label{sec:treebanks}
\label{sec:data:pos}

In order for our parsers to work well with predicted POS tags, 
we follow the same steps as used in the 2018 CoNLL shared task 
for creating training and development treebanks with automatically predicted POS tags (henceforth referred to as silver POS).
Since we are  required to parse translated text which only has lexical features available, we disregard
lemmas, language-specific POS (XPOS) and morphological features and only use the word form and universal POS (UPOS) tag as input features to our parsers.
We develop our POS tagging and parsing models using the AllenNLP library \cite{gardner-etal-2018-allennlp}.

We use jackknife resampling to predict the UPOS tags for the training treebanks.
We split the training treebank into ten parts, train models on nine parts and predict UPOS for the excluded part.
The process is repeated until all ten parts are predicted and they are then combined to recreate the treebank with silver POS tags.
Only token features are used to predict the UPOS tag.\footnote{We
    observe slightly lower POS tagging scores on fully annotated test sets than
    UDPipe, which uses gold lemmas, XPOS and
    morphological features to predict the UPOS
    label and therefore cannot be applied to
    the translated text without also building
    predictors for these features.
}
Finally, we  train a model per source language on the full training data to check performance on the respective development set and to POS tag the source language translations before parsing.

\begin{figure}[htb]
\centering\includegraphics[width=7.62cm]{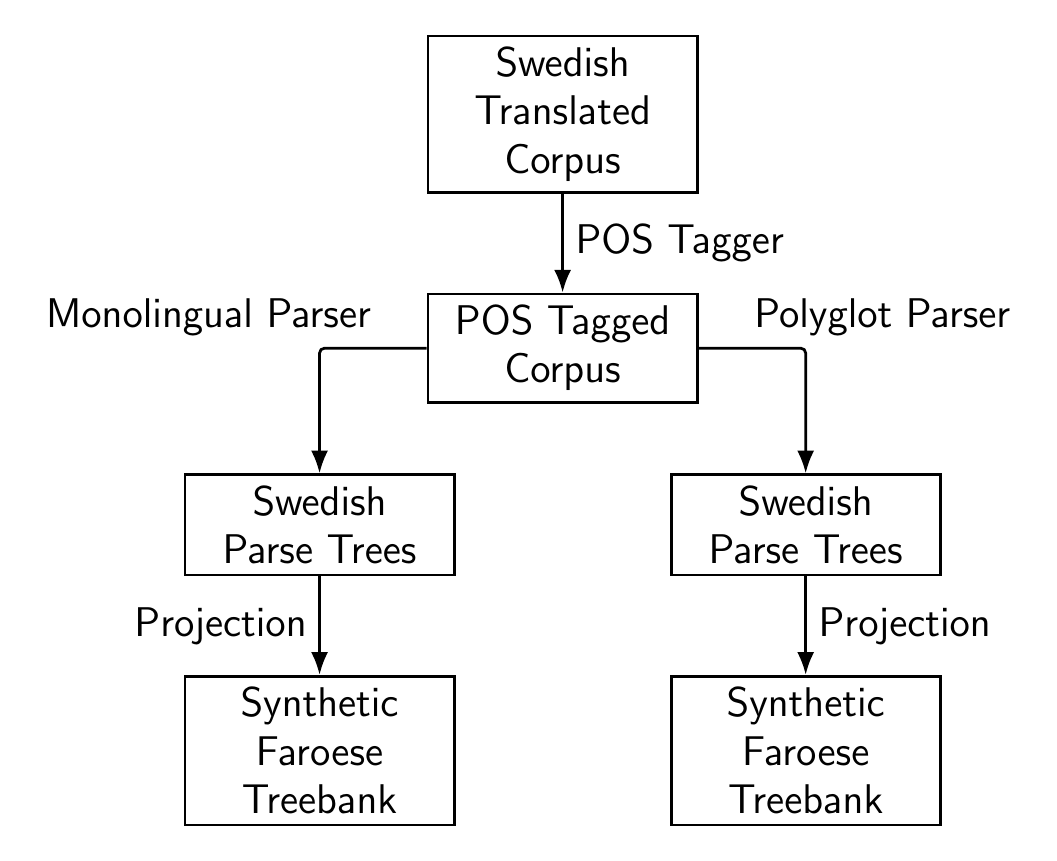}
\caption{Overview of the \textit{monolingual} and \textit{polyglot} parse experiments using Swedish translations as an example. This process is repeated for all source languages.}
\label{fig:overview}
\end{figure}

We train two variants of parsing models.
The first is a monolingual biaffine dependency parser \cite{dozat-manning} trained on the individual source treebanks. The second is a polyglot model trained on
all source treebanks using the multilingual parser of \newcite{schuster-etal-2019-cross}, which is the same graph-based biaffine dependency parser, extended to enable parsing with multiple treebanks.
We additionally include a treebank embedding \cite{ammar-etal-2016-many, P18:2098} to the input of the polyglot parser 
to help the parser differentiate between the source languages. We optimize the model for average development set LAS across the included languages.
The process is illustrated in Fig.~\ref{fig:overview}.

To ensure that our parser is realistic, we add
a pre-trained monolingual word embedding to each monolingual parser, giving a considerable improvement in 
accuracy on the development sets of the source languages.
We use the precomputed Word2Vec embeddings\footnote{\url{https://lindat.mff.cuni.cz/repository/xmlui/handle/11234/1-1989}} released as part of the 2017 CoNLL shared task on UD parsing \cite{zeman:EtAl:2017:K17:3} which were trained on CommonCrawl and Wikipedia.

In order to use pre-trained word embeddings for the polyglot setting, we need to consider that
a polyglot model uses a shared vocabulary across all input languages.
In our experiments, we simply use the union of the word embeddings and average the word vector for words that occur in more than one language.
Future work should explore cross-lingual word embeddings with limited amount of parallel data or use aligned contextual embeddings as in \cite{schuster-etal-2019-cross}.

\paragraph{Synthetic Source Treebanks}
\label{sec:method:sst}

Source translations are tokenized with UDPipe~\cite{straka-strakova-2017-tokenizing} by \newcite{tyers-etal-2018-multi}.
For each source language, the POS model trained on the full training data (see previous section) is used
to tag the tokenized translations. 
Once the text is tagged, we predict dependency arcs and labels with the parsing models of the previous section, and use annotation projection (described below) to provide syntactic annotations for the target sentences.

\paragraph{Annotation Projection}
\label{sec:method:projection}
Once the
synthetic source treebanks are compiled, \ie the translations are parsed,
the annotations are then projected from the source translations to the target language using the word alignments and \citeauthor{tyers-etal-2018-multi}'s projection tool, resulting in a Faroese treebank.
In some cases, not all tokens are aligned and \newcite{tyers-etal-2018-multi} work around this by falling back to a 1:1 mapping between the target index and the source index.
There are also cases where there is a mismatch in length between the source and target sentences and some dependency structures cannot be projected to the target language.
\citeauthor{tyers-etal-2018-multi}'s projection setup removes unsuitable projected trees containing \eg more than one root token, a token that is its own head or a token with a head outside the range of the sentence.

\paragraph{Multi-source Projection}
\label{sec:multi-source}

\begin{figure*}[ht]
\centering\includegraphics[width=14cm]{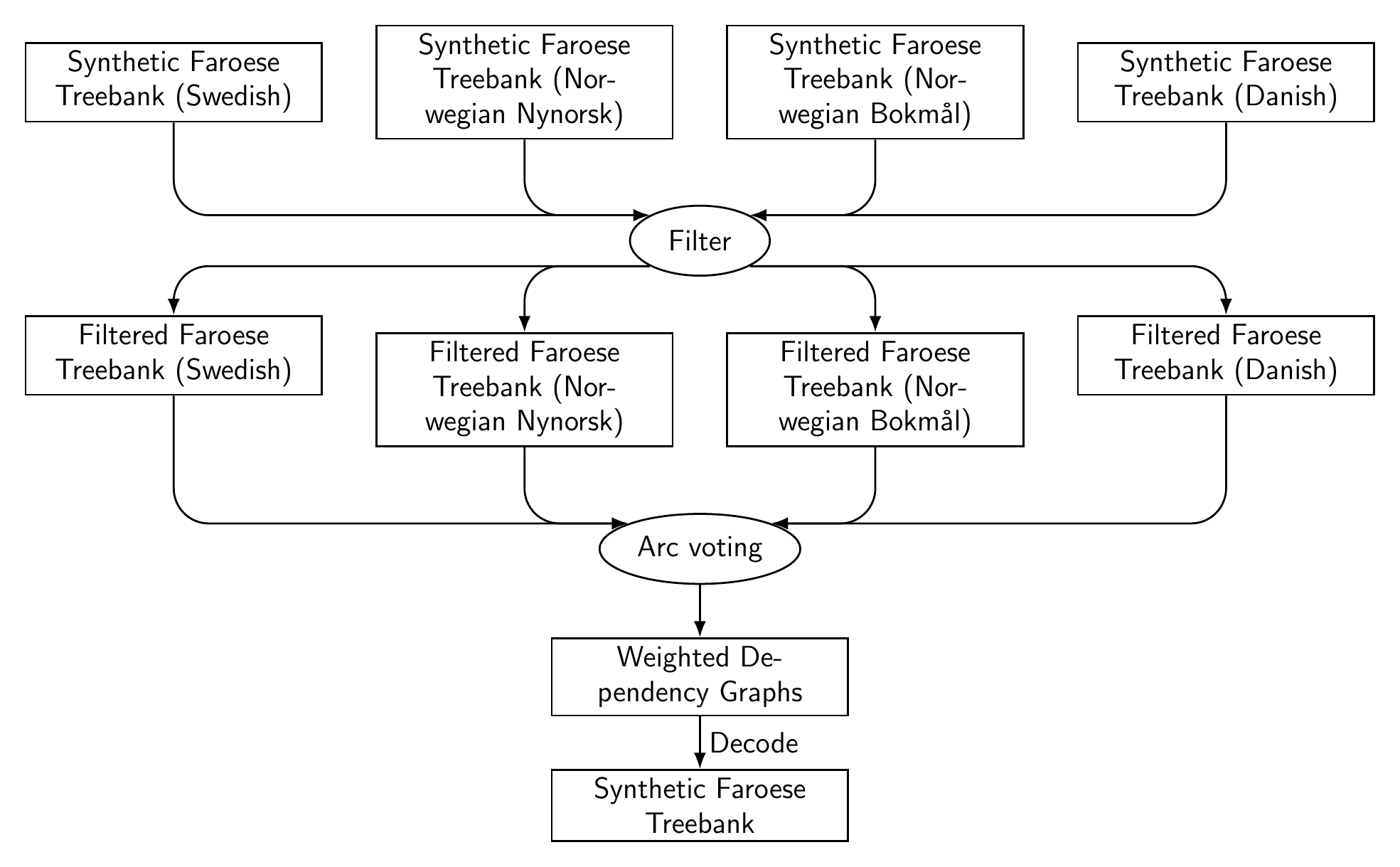}
\caption{Multi-source projection. The source language is listed in brackets.}
\label{fig:msp}
\end{figure*}

For multi-source projection, the four source-language dependency trees for a Faroese sentence are projected into a single graph, scoring edges according to the number of trees that contain them \cite{sagae-lavie-2006-parser, nivre-etal-2007-conll}.
The dependency structure is first built by voting over the directed edges. Afterward, dependency labels and POS tags are decided using the same voting procedure.
The process is illustrated in Fig.~\ref{fig:msp}.

\paragraph{Target Tagging and Parsing Models}

At this stage we have Faroese treebanks to train our 
POS tagging
and parsing models.
The Faroese treebanks come in two variants: the result of projection from source trees produced by either 1) a monolingual, or 2) the polyglot model.
For each case, we train our POS tagging and parsing models directly on these synthetic treebanks and do not make use of word embeddings as we do not have them for Faroese.

\begin{figure*}[ht]
    \centering\includegraphics[width=14.5cm]{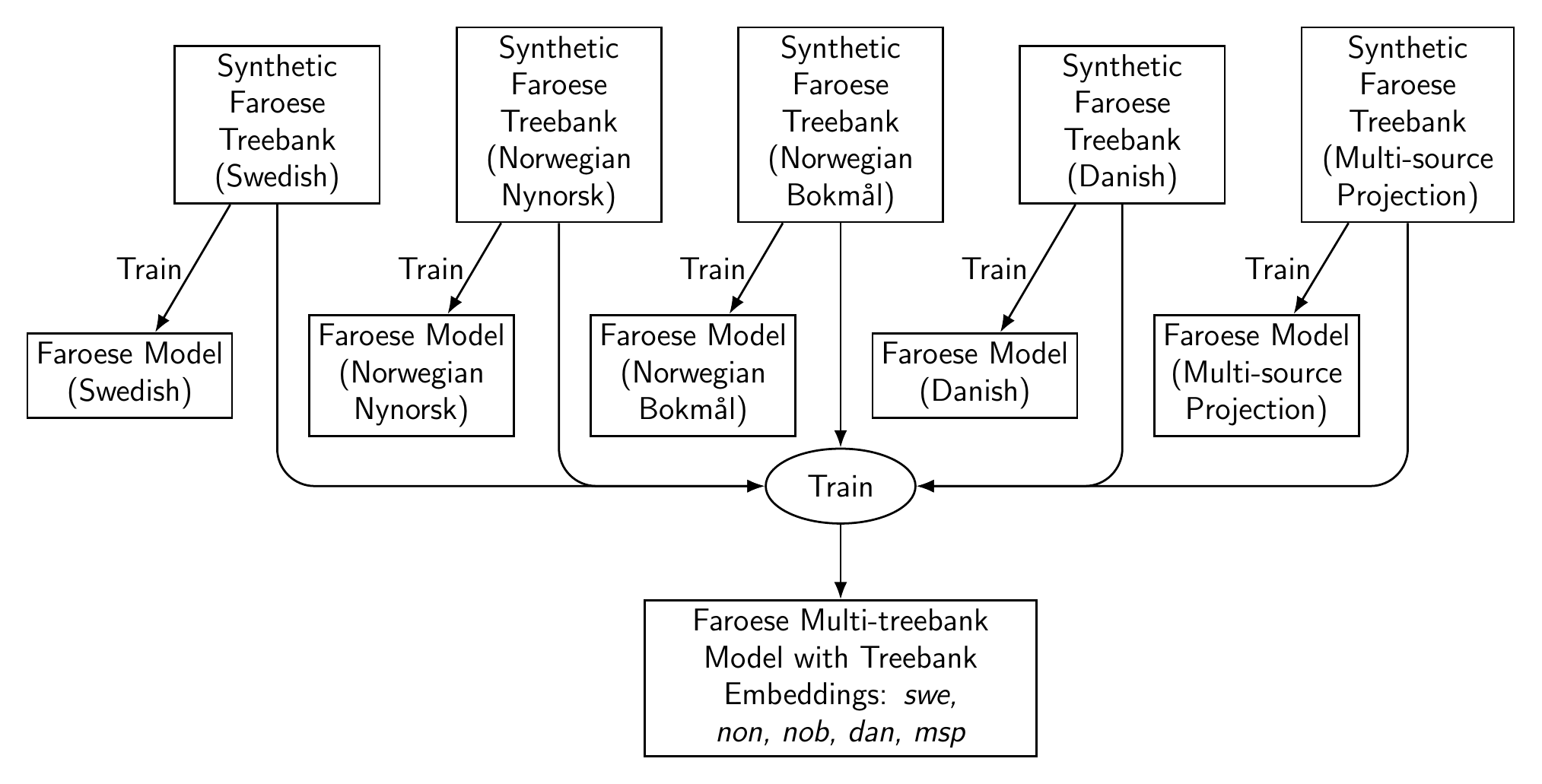}
    \caption{Single versus multi-treebank training. The source language is listed in brackets.}
    \label{fig:multi-tb}
\end{figure*}

\paragraph{Multi-treebank Target Parsing}
Since we have several synthetic Faroese treebanks, we have the option of training on a single  treebank or using a multi-treebank approach where we train on all target treebanks in the same way as we did for inducing the polyglot source model.
The process is illustrated in Fig.~\ref{fig:multi-tb}.
When training a multi-treebank target model, for each target treebank, we add a treebank embedding denoting the source model used to project annotations to the target treebank.
At predict time, we must include one of these treebank embeddings as input to the model.
As we do not have real Faroese data in our target training treebanks, we must choose the treebank embedding of one of the synthetic target treebanks.
\newcite{P18:2098} introduce the term ``proxy treebank'' to refer to cases where the test treebank is not in the training set and a treebank embedding from the training set must be used instead.

%% file: model-details.tex

\begin{table}[t!]
\begin{center}
\small
\begin{tabular}{|l|l|}
\hline
\multicolumn{2}{|c|}{\bf POS Tagger Architecture } \\ \hline
\bf Parameter        & \bf Value \\ \hline
Char-BiLSTM layers & 2 \\
BiLSTM layers      & 2 \\
BiLSTM size        & 400 \\
Dropout LSTMs      & 0.33 \\
Dropout MLP        & 0.33 \\
Dropout embeddings & 0.33 \\
Nonlinear act. (MLP) & ELU \\
\hline
\hline
\multicolumn{2}{|c|}{\bf Parser Architecture } \\ \hline
\bf Parameter        & \bf Value \\ \hline
Char-BiLSTM layers & 2 \\
BiLSTM layers      & 3 \\
BiLSTM size        & 400 \\
Arc MLP size     & 500 \\
Label MLP size & 100 \\
Dropout LSTMs      & 0.33 \\
Dropout MLP        & 0.33 \\
Dropout embeddings & 0.33 \\
Nonlinear act. (MLP) & ELU \\
\hline
\hline
\multicolumn{2}{|c|}{\bf Embeddings } \\ \hline
\bf Parameter &\bf Size \\
Word embedding (both)       & 100 \\
Char embedding (both)       & 64 \\
POS embedding (parser)       & 50 \\
Treebank embedding (both)  & 12 \\
\hline
\hline
\multicolumn{2}{|c|}{\bf Training } \\ \hline
\bf{Parameter}        & \bf Value \\ \hline
Optimizer          & Adam\\
Learning rate      & 0.001\\
Adam epsilon       & 1e-08\\
beta1 (both)             & 0.9\\
beta2 (parser)              & 0.9\\
beta2 (tagger)             & 0.999\\
\hline
\end{tabular}
\end{center}
\caption{\label{table:hyper-params} Chosen hyperparameters for our POS tagging and parsing models. \emph{both} means the feature is common to both the POS tagger and parser.}
\end{table}

%% file: experiments.tex
\label{sec:experiments}

The hyper-parameters of our POS tagging and parsing models are given in Table~\ref{table:hyper-params}.
For POS tagging, we adopt a standard architecture with a word and character-level Bi-LSTM \cite{plank-etal-2016-multilingual,graves2005framewise} to learn context-sensitive representations of our words.
These representations are passed to a multilayer perceptron (MLP) classifier followed by a softmax function to choose a tag with the highest probability.
For both the POS tagging and parsing models, we use a word embedding dimension of size 100 and a character embedding dimension of size 64.
POS tag embeddings of dimension 50 are included in the parser.
We train our Faroese models for fifty epochs.
We do not split the synthetic Faroese treebanks into training/development portions though we suspect doing so will help the models to not overfit on the training data.
For all experiments we report labelled attachment scores produced by the official CoNLL 2018 evaluation
script.\footnote{\url{https://github.com/ufal/conll2018/blob/master/evaluation_script/conll18_ud_eval.py}}

%% file: results_dev.tex
\begin{table}
\small
\begin{center}
\begin{tabular}{ ccc }
\toprule
{\sc Treebank} & {\sc Monolingual} & {\sc Polyglot} \\
\midrule
da\textunderscore{ddt}       &     81.10 & \bf 82.75 \\ 
sv\textunderscore{talbanken} &     80.61 & \bf 83.85 \\
no\textunderscore{nynorsk}   & \bf 88.54 &     88.29 \\
no\textunderscore{bokmaal}   &     89.29 & \bf 90.29 \\
\midrule
average                      &     84.88 & \bf 86.30 \\
\bottomrule
\end{tabular}
\caption{Source model LAS scores on the development treebanks using silver POS tags.}
\label{tab:dev_results}
\end{center}
\end{table}

%% file: sentence-stats.tex

\begin{table}[t!]
\small
\begin{center}
\setlength{\tabcolsep}{3.0pt}
\begin{tabular}{lcc}
\toprule
\sc Source         & \sc Monolingual & \sc Polyglot \\
\midrule
Danish               &  13,950 &  13,944 \\
Swedish              &  10,894 &  10,874 \\
Norwegian Nynorsk    &  13,177 &  13,194 \\
Norwegian \bokmaal{} &  17,345 &  17,378 \\
Multi-source         &   6,716 &   6,833 \\
\bottomrule
\end{tabular}
\end{center}
\caption{The number of valid sentences in the Faroese synthetic treebank for each source language after annotation projection and sentence filtering. }
\label{tab:target-sentences}
\end{table}

%% file: results_faroese.tex
\begin{table}[t!]
\centering
\small
\begin{tabular}{llcc}
\toprule
\sc Source Language & \sc Source & \multicolumn{2}{c}{\sc Target Model } \\
 & \sc Model &  \sc Single &  \sc Multi  \\
\midrule
\multirow{2}{*}{Danish}
        & Monolingual & 61.24 &     63.40 \\%

        & Polyglot    & 65.29$^\dagger$  & \bf 65.53$^\dagger$  \\%
\addlinespace
\multirow{2}{*}{Swedish}
        & Monolingual & 65.93 &     66.15 \\%

        & Polyglot    & 68.60$^\dagger$  & \bf 69.69$^\dagger$  \\%
\addlinespace
\multirow{2}{*}{Norwegian Nynorsk}
        & Monolingual & 70.27 & \bf 71.51 \\%

        & Polyglot    & 69.80 &      71.13 \\%
\addlinespace
\multirow{2}{*}{Norwegian \bokmaal{}}
        & Monolingual & 67.46 &     67.94 \\%

        & Polyglot    & 70.51$^\dagger$  & \bf 70.58$^\dagger$  \\%
\addlinespace
\multirow{2}{*}{Multi-source}
        & Monolingual & 68.00 & 69.80 \\%

        & Polyglot    & 68.55 & \bf 70.07 \\%
\addlinespace
\hline
\addlinespace
\multirow{2}{*}{ Average}
        & Monolingual & 66.58 &     67.76 \\%

        & Polyglot    & 68.55 & \bf 69.40 \\%
\addlinespace
\bottomrule
\end{tabular}
\caption{\label{table:results-faroese} LAS on the target Faroese test treebank.
\emph{Single} refers to using a single synthetic Faroese treebank to train a Faroese model,  \emph{Multi} uses both a multi-treebank POS tagger and a multi-treebank parser with all synthetic Faroese treebanks.
The multi-treebank model is tested with each of the five training treebanks (four projected from individual source languages and one using multi-source projection) as proxy treebank.
Statistically significant differences 
between the monolingual and polyglot setting are indicated by $\dagger$ for each result pair, excluding averages.
}
\end{table}

%% file: results-sampled.tex
\begin{table}[htb]
\centering
\small
\begin{tabular}{*{3}{l}}
\toprule
\sc Source Language & \sc Monolingual & \sc Polyglot  \\
\midrule
Danish               &  61.13 &  \bf 64.43 \\
Swedish              &  63.19 &  \bf 67.46 \\
Norwegian Nynorsk    &  68.72 &  \bf 69.28 \\
Norwegian \bokmaal{} &  66.13 &  \bf 68.77 \\
Multi-source         &  68.00 &  \bf 68.55 \\
\midrule
Average              &  65.43 &  \bf 67.70 \\
\bottomrule
\end{tabular}
\caption{\label{table:sampled} LAS scores between target models trained on the subset of sentences eligible for multi-source projection (with annotations from the stated source).}
\end{table}

%% file: results-previous.tex
\begin{table}[htb]
\centering
\begin{tabular}{*{3}{l}}
\toprule
\sc Work & \sc Result  \\
\midrule
\newcite{rosa-marecek-2018-cuni} & 49.4 \\
\hline
\newcite{tyers-etal-2018-multi} &  64.4\\
Our implementation &  68.0 \\
of \newcite{tyers-etal-2018-multi} & & \\
Our Best Model & \bf 71.5 \\
\bottomrule
\end{tabular}
\caption{\label{table:previous} Comparison to previous work. LAS on Faroese test set. Note that the first results uses predicted segmentation and tokenization whereas the rest used gold.}
\end{table}

%% file: results.tex
The development results of our monolingual and polyglot models on the source language treebanks are shown in Table \ref{tab:dev_results}.
The results for the polyglot model are better for three out of four source languages, whereas for no\_nynorsk, the monolingual model marginally outperforms the polyglot one.
These results suggest that the polyglot model will contribute better syntactic annotations for Faroese treebanks. 

The statistics of the filtered Faroese treebanks obtained via projection with our source parsing models 
are given in Table~\ref{tab:target-sentences}.
The treebank sizes are fairly similar regardless of whether source annotations are provided by a monolingual or a polyglot model
which is expected because the word alignments are the major factor in determining whether a projection is successful.
There is a proportionally lower number of sentences for multi-source projection.
This is because this method only uses the intersection of sentences which are present across all synthetic treebanks after filtering.
The treebank originating from Norwegian \bokmaal{} has the highest number of valid sentences, suggesting that it could be a good candidate for projection to Faroese. It also has the highest source language parsing accuracy (Table~\ref{tab:dev_results}).

The results of training on our various synthetic Faroese  treebanks and predicting the Faroese test set are shown in the first result column of Table~\ref{table:results-faroese} (\texttt{SINGLE}).
In terms of monolingual \vs polyglot, we find that projecting from a polyglot model helps with four out of the five possible treebanks (with three of them being statistically significant).\footnote{Statistical significance is tested with udapi-python \url{https://github.com/udapi/udapi-python}.
LAS differences are reported as significant if p $<$ 0.05.}
The polyglot model was
outperformed by
the monolingual model using Norwegian Nynorsk for
projection though the difference is not statistically significant.
On the source side, the monolingual Norwegian Nynorsk model also performed slightly better than the polyglot model (Table~\ref{tab:dev_results}).
This observation supports the intuition that the quality of the projected annotations can be improved by contributing better source annotations, \ie improving the source model(s) is one way to improve performance of the target model. This is supported by the fact that the source language with the highest LAS (Norwegian \bokmaal{}) is also the best choice for projection (in this single target model setting).

The multi-source approach was not that effective in our case and  some individual better sources were able to surpass this combination approach.
One could argue that this may be due to the lower amount of training data when using the multi-source treebank.
We test this hypothesis by
only including those sentences which contributed to multi-source projection in the single-source synthetic treebanks.
The results are given in Table~\ref{table:sampled}. Comparing the results in Tables~\ref{table:results-faroese} and \ref{table:sampled}, we see that LAS scores tend to be slightly lower than on the version which included all target sentences, indicating that we did lose some information by filtering out a large number of sentences. However, Norwegian Nynorsk still outperforms the multi-source model for the monolingual setting and both Norwegian models perform better than the multi-source model in the polyglot setting, suggesting that size alone does not explain the under-performance of the multi-source model.
It is also worth noting that polyglot training is superior to all monolingual models which hints that for no\_nynorsk (the previously better performing model), the monolingual model was not able to achieve its full potential with the reduced data while the polyglot model was able to provide richer annotations.

Another reason why the multi-source model does not work as well in our experiments as it does in those of \newcite{tyers-etal-2018-multi} might be that we use pre-trained embeddings whereas \newcite{tyers-etal-2018-multi} do not.
In this way, our monolingual models are stronger and likely do not benefit as much from voting.

The second result column (\texttt{MULTI}) of
Table~\ref{table:results-faroese}
shows the effect of 
training a multi-treebank POS tagger and parser on the Faroese treebanks created by each of the four source languages as well as the treebank which is produced by multi-source projection.
This experiment is orthogonal to the experiment using a polyglot model on the source side and so we also test a combination of polyglot source side parsing and multi-treebank target side parsing.
We see improvements over the single treebank setting 
for all cases.\footnote{The multi-treebank tagger closely resembles the dependency parser, where we add a treebank embedding and optimize for average accuracy across the included treebanks.
To the best of our knowledge, this is the first reported use of a multi-treebank POS tagger using a treebank embedding~\cite{P18:2098}. We also tested the effect of training only the dependency parser using multiple treebanks but found that it always helps to also perform multi-treebank training for the POS tagger.}

Table~\ref{table:previous} places our systems in the context of previous results on the same Faroese test set.
The highest scoring system in the 2018 CoNLL shared task was that of \newcite{rosa-marecek-2018-cuni} who achieved a LAS score of 49.4 on the Faroese test set.
Note that they use predicted tokenization and segmentation whereas our experiments and \citeauthor{tyers-etal-2018-multi}'s use gold tokenization and segmentation, which provides a small  artificial boost. 
\newcite{tyers-etal-2018-multi} report an LAS of 64.43 with a monolingual multi-source approach. Our implementation which uses a different parser (AllenNLP versus UDPipe) and pre-trained word embeddings achieves an LAS of 68. 
Our highest score of 71.51 is achieved through the combination of projecting from strong monolingual source models and then training multi-treebank POS tagging and parsing models on the outputs.

%% file: conclusion.tex
We have presented parsing results on Faroese, a low-resource language, using annotation projection from multiple monolingual sources versus a single polyglot model.
We also extended the idea of multi-treebank learning to the target treebanks.

The results of our experiments show that the use of a polyglot source model helps in four out of five cases 
using  single treebank target models.
The two source languages that have lowest LAS when using monolingual parsers, namely Danish and Swedish, see
significant improvements
when switching to a polyglot model.
Our best performing single target model
is trained on Faroese trees projected from 
Norwegian \bokmaal{} trees produced by a polyglot model.
However, the strongest language with
monolingual modelling, Norwegian Nynorsk, does not benefit
from switching to a polyglot model.
When we filtered the target treebank to the subset of sentences selected by multi-source projection, the polyglot model is superior to all five monolingual models, even outperforming the Norwegian Nynorsk model.
One explanation of the improvements seen with polyglot modelling is that it introduces a new interaction point for cross-lingual features via the feature extractor of the polyglot parser.
With monolingual source models, cross-lingual features only interact indirectly in the graph-decoding stage of multi-source projection.

We also applied the multi-treebank approach to the target-side POS tagger and parser and 
see improvements for all settings.
The overall best result
is with the setting that uses monolingual source models to create the source trees that are projected to Faroese and combined in a multi-treebank model.
The proxy treebank for the multi-treebank model is the treebank that also gave best results with single treebank target models, projected from Norwegian Nynorsk.

We presented a simple solution to deal with using multiple pre-trained embeddings in a model with a shared vocabulary. 
It was a rather \naive{} solution and 
we want to explore the use of available cross-lingual word embedding tools.
Additionally, the use of contextual embeddings such as ELMo \cite{peters-EtAl:2018:N18-1} or multilingual BERT \cite{devlin-etal-2019-bert} would likely provide better representations, with the effect of contributing better annotations for the target language.
Indeed, recent work has already shown promising work in this area \cite{schuster-etal-2019-cross, DBLP:journals/corr/abs-1904-02099}.

In the multi-source projection experiments, our criteria for filtering is based on whether the sentence was present across all target treebanks and more sophisticated approaches could be used to select better training instances as in \newcite{plank-agic-2018-distant}.

More generally, we would like to investigate how our findings might change when the number of source languages or treebanks is changed and how the observations carry over to other languages than Faroese. 
It would also be interesting to use multiple sources of arc weights in a dense graph as in \cite{agic-etal-2016-multilingual} but with models induced from training on multiple source languages together.
To work with language pairs with more deviating word orders and/or translations that are not word-for-word,
the choice of word alignment algorithm and the
projection algorithm may have to be revised.

%% file: ack.tex
This  research  is  supported  by  Science  Foundation Ireland
through the ADAPT Centre for Digital Content Technology, which is
funded under the SFI Research Centres Programme (Grant 13/RC/2106)
and is co-funded under the European Regional Development Fund.
We thank
\newcite{tyers-etal-2018-multi} 
for releasing their 
data and
the anonymous reviewers for their helpful feedback.